\pdfoutput=1
\documentclass[letterpaper,10pt,conference]{ieeeconf}
\IEEEoverridecommandlockouts
\usepackage{amsmath}
\usepackage{amssymb}

\usepackage{graphicx}
\usepackage{subcaption}
\usepackage{caption}
\usepackage{booktabs}
\usepackage{multirow}
\usepackage{array}
\usepackage{tabularx}
\usepackage{placeins}
\usepackage{arydshln}
\usepackage{float}
\usepackage{xcolor}
\usepackage[T1]{fontenc}
\usepackage[utf8]{inputenc}
\usepackage{xspace}
\usepackage{url}
\usepackage{cite}
\usepackage[hidelinks]{hyperref}
\hypersetup{
  pdftitle={SPINE: Bridging the Cyber-Physical Gap with Agentic AI},
  pdfauthor={Minkyu Ham, Dongho Kim, Chan Lee, Min Jun Kim, Yixi Zhang, Jiayi Wang, Han Liu}
}

\title{\LARGE \bf
SPINE: Bridging the Cyber-Physical Gap with Agentic AI}

\author{
Minkyu Ham$^{1,*}$, Dongho Kim$^{1,*}$, Chan Lee$^{1,*}$, Min Jun Kim$^{1}$, Yixi Zhang$^{2}$, Jiayi Wang$^{2}$, and Han Liu$^{1,2,\dagger}$\\[0.4em]
\small Website: \href{https://magics-lab.github.io/SPINE-web/}{\texttt{magics-lab.github.io/SPINE-web}}%
\thanks{$^{*}$These authors contributed equally to this work and are ordered alphabetically by last name.
$^{1}$Department of Statistics and Data Science, Northwestern University.
$^{2}$Department of Computer Science, Northwestern University.
$^{\dagger}$Correspondence: \texttt{hanliu@northwestern.edu}.}%
}

\begin{document}
\maketitle

\begin{abstract}
Foundation models give robots powerful high-level reasoning, yet turning that intelligence into reliable physical action remains difficult: roboticists must still align device drivers, network interfaces, sensors, controllers, and safety constraints for each platform. This often-overlooked integration layer acts as the robot's spinal cord, translating high-level intent into coordinated physical behavior, and remains a primary bottleneck for scalable Embodied AI. Hence, we propose SPINE (Scalable Physical Integration with ageNtic Expertise), an agentic framework for systematically debugging and deploying bimanual robots for teleoperation. SPINE centers on two subagent-driven workflows: a profile builder that compiles robot-specific context and a debugger that uses that context to iterate through diagnosis, repair, and validation until teleoperation succeeds. Across two bimanual robot platforms and 12 debugging scenarios, novice-operated SPINE achieved more complete and efficient recovery than human operators using Claude Code. On DOBOT X-Trainer, SPINE improved success from 76\% to 100\% and reduced mean time-to-teleoperation by 30\%; on AgileX PiPER, SPINE also achieved 100\% success and reduced mean time-to-teleoperation by 38\%. These results show that structured agentic debugging can address a key cyber-physical integration bottleneck in real-world robot deployment.
\end{abstract}

\section{Introduction}\label{sec:intro}

Over the past decade, robot learning has expanded robots' ability to acquire skills from data and reuse experience across tasks and platforms. Deep visuomotor policies demonstrated learned mappings from camera images to motor commands \cite{Levine2016Visuomotor}, while Diffusion Policy and RT-1 extended learned manipulation through diffusion-based action generation and transformer policies trained on diverse demonstrations \cite{Chi2023DiffusionPolicy,Brohan2022RT1}. Open X-Embodiment further demonstrated the benefits of sharing training data across robotic platforms \cite{Padalkar2023}. Foundation models have accelerated these advances by bringing visual and linguistic knowledge into planning and control. SayCan grounds language-based plans in the skills a robot can execute, while Code as Policies translates instructions into programs that combine perception and control functions \cite{Ahn2022,Liang2022CodeAsPolicies}. Vision-language-action models, including RT-2, OpenVLA, and $\pi_0$, combine pretrained representations with robot demonstrations to generate actions from observations and instructions \cite{Zitkovich2023RT2,Kim2024OpenVLA,Black2025Pi0}. Together, these advances broaden robot capabilities and support generalization to new tasks, objects, and environments.

Deploying these models still requires a functioning software, communication, and hardware stack, yet dependency conflicts, configuration errors, and component incompatibilities can prevent basic operation \cite{Timperley2024ROBUST,Canelas2024Misconfigurations}. Debugging this integration layer demands broad expertise because software and hardware faults can produce similar symptoms and persist in combination. A missing camera stream, for example, can reflect an incorrect device identifier, an incompatible driver, or a disconnected cable. If the configuration and cable are both faulty, correcting either one alone leaves the stream unavailable. Operators must therefore connect evidence from source code, configuration files, runtime logs, and physical inspection, then verify whether their interventions restore the complete system. Studies of robotics development identify substantial integration effort and platform-specific variation across electronics, mechanics, communication protocols, and software \cite{Garcia2023Variability}. These demands create an expertise barrier for newcomers and can divert research effort from experimentation toward establishing basic operability.

\begin{figure}[t]
\centering
\begin{subfigure}[t]{0.48\columnwidth}
  \centering
  \includegraphics[width=\linewidth, height=3.5cm]{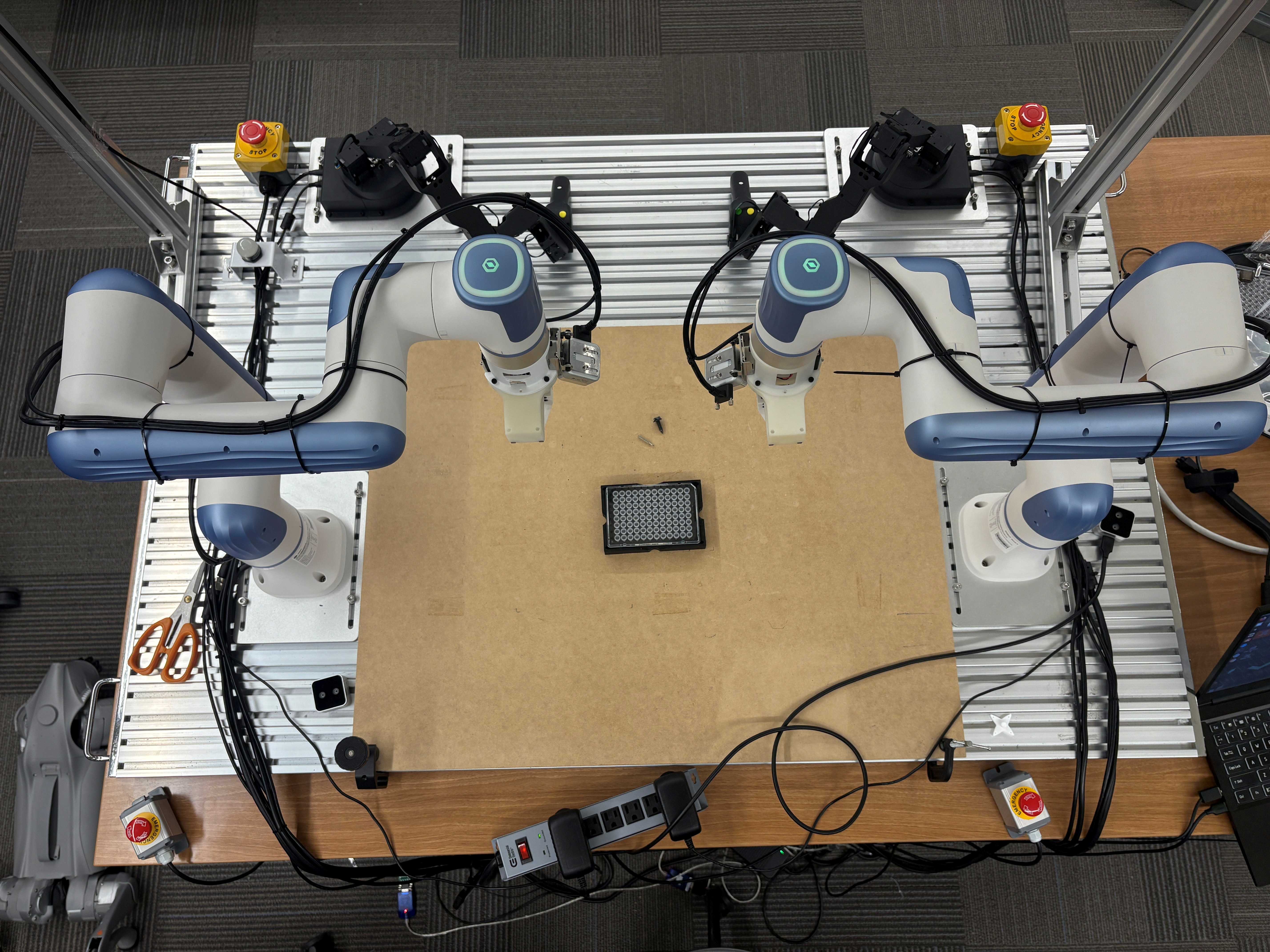}
  \caption{DOBOT X-Trainer}
\end{subfigure}\hfill
\begin{subfigure}[t]{0.48\columnwidth}
  \centering
  \includegraphics[width=\linewidth, height=3.5cm]{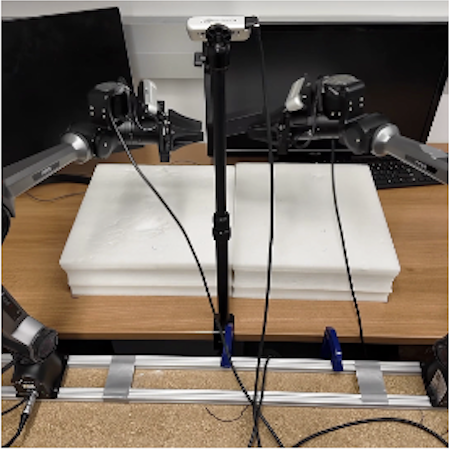}
  \caption{AgileX PiPER}
\end{subfigure}
\caption{Bimanual robot platforms used in the evaluation.}
\label{fig:platforms}
\end{figure}

Existing recovery and diagnostic methods address parts of this problem, but their evaluations do not establish complete recovery from combined software, communication, and physical faults. DROC and RACER address task-execution failures, while ASPIRE debugs control programs through existing robot interfaces \cite{Zha2024DROC,Dai2025RACER,Lu2026ASPIRE}. Operators can also use LLM-based coding agents to inspect and repair software \cite{Yang2024,AnthropicClaudeCode}, but code edits alone cannot reconnect hardware or establish that middleware communicates with the intended devices. The remaining gap is coordinating software repair, live middleware and device checks, and guided physical interventions until validation confirms that the complete robot operates.

Closing this deployment gap requires a coordinated process that gathers evidence across software and hardware, selects digital or physical interventions, and verifies recovery through live execution. In this work, \emph{operational} specifically means \emph{teleoperation-ready}: an operator can control the follower arms and grippers through the master arms while viewing the required camera streams. This definition does not imply readiness for autonomous policy execution. Making this process systematic and accessible would reduce the expertise required to establish the working physical systems that robot learning depends on.

\begin{figure*}[t]
\centering
\includegraphics[width=0.98\textwidth]{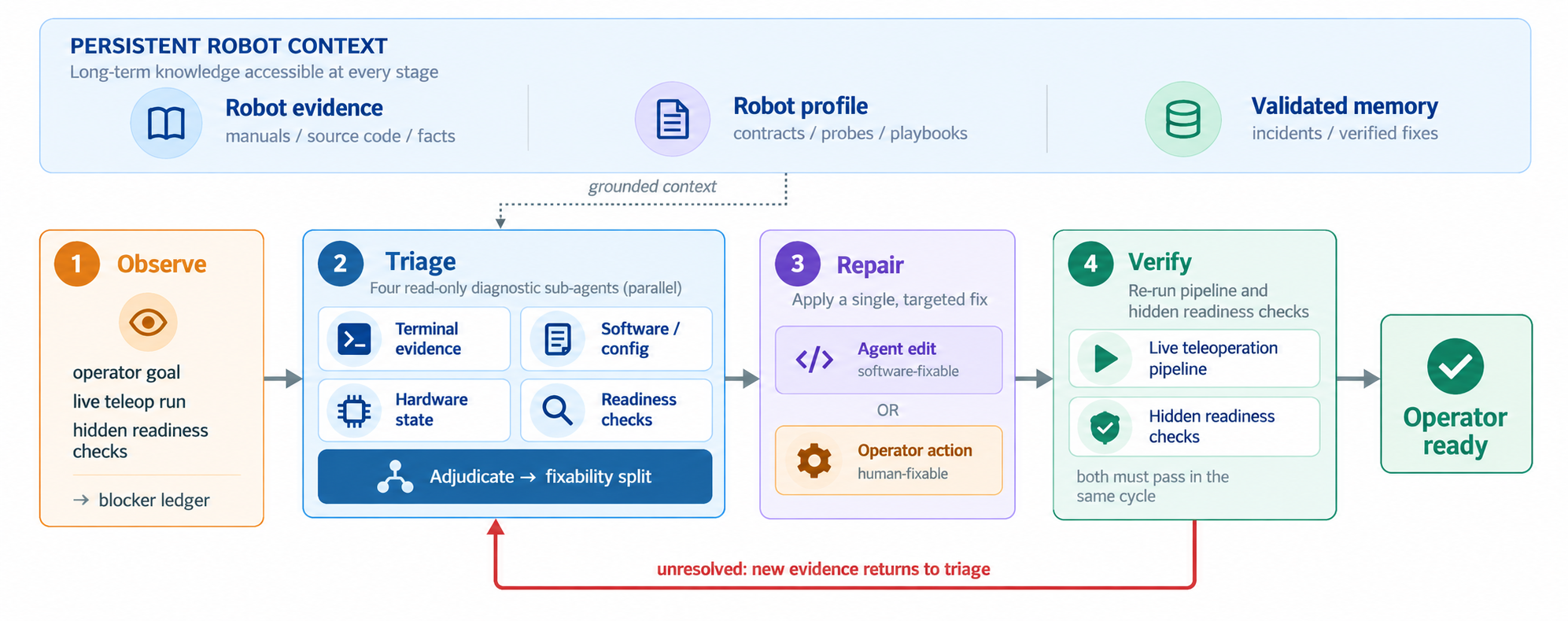}
\caption{SPINE Debugging Workflow: separates persistent robot knowledge from a goal-conditioned recovery loop. Triage dispatches four read-only diagnostic sub-agents in one parallel call and adjudicates their reports into a fixability split; repair is a separate stage that applies either a software edit or one operator action. Verification reruns the live teleoperation pipeline and hidden readiness checks together, and unresolved cases return to triage with new evidence rather than restarting observation.}
\label{fig:architecture}
\end{figure*}

To address this gap, we introduce SPINE (Scalable Physical Integration with ageNtic Expertise), an agentic harness for restoring teleoperation with minimal robotics expertise. SPINE runs within Claude Code \cite{AnthropicClaudeCode} and centers on two subagent-driven workflows: a profile builder and a debugger (Fig.~\ref{fig:architecture}). The profile builder compiles documentation, known-good source code, and operator-confirmed facts into a persistent reference for diagnosis, repair, and validation. The debugger combines this reference with live evidence and incident memory to coordinate parallel diagnosis, targeted software repairs, and step-by-step guidance for physical interventions. It iterates through diagnosis, repair, and validation, closing a case only when the complete teleoperation pipeline and profile-defined readiness checks pass in the same validation cycle. SPINE retains incident outcomes and validation evidence to inform subsequent debugging sessions.

We evaluate novice-operated SPINE across twelve software-only, hardware-only, and hybrid debugging scenarios on DOBOT X-Trainer and AgileX PiPER. We compare it with human operators using Claude Code and the same underlying language model, but without SPINE's structured workflow. SPINE resolved all implanted faults on both platforms (Table~\ref{tab:scenario_results}). On DOBOT, it increased the mean fraction of faults resolved from 76\% to 100\% and reduced mean time-to-teleoperation by 30\%; on PiPER, it increased fault resolution from 99\% to 100\% and reduced mean time-to-teleoperation by 38\%. These results demonstrate more complete and efficient teleoperation recovery across the two platforms, including scenarios that require both software repair and physical intervention.

DOBOT ablations suggest that both the robot profile and hardware playbook contribute to SPINE's performance (Table~\ref{tab:dobot_ablation_results}). Removing them individually reduced mean fault resolution from 100\% to 79\% and 57\%, respectively, and increased mean debugging duration. With only one trial per scenario for each ablated variant, these findings provide preliminary support for robot-specific context and structured physical guidance.

Together, these results support a role for agentic AI beyond controlling already functional robots: helping operators make them teleoperation-ready. SPINE advances this direction through structured, validated debugging that bridges software and hardware, making teleoperation recovery more accessible and efficient.

\section{Related Work}\label{sec:related}
SPINE is closest to work on robot failure recovery, language-guided robot systems, and tool-using software agents. The distinction is the starting state: SPINE assumes that the robot may not yet expose a reliable action interface, so recovery must span source, middleware, devices, and operator-visible hardware before task execution can resume.

\subsection{Robot deployment, fault diagnosis, and recovery}
ROS and ROS~2 standardize communication among robot drivers, sensors, controllers, and applications \cite{Quigley2009, Macenski2022}, but deployment state remains distributed across launch files, environment variables, device identifiers, network settings, and vendor libraries. Industrial robotics studies likewise identify integration, maintenance, heterogeneous interfaces, and fault recovery as persistent barriers to dependable operation \cite{Urrea2025IndustrialRobotics}. These systems provide the substrate on which SPINE operates, but they do not assemble platform-specific evidence into an end-to-end recovery decision.

Systems work also reduces particular deployment costs. SARA-RT targets the computational efficiency of large robot transformers, and Robo-DM targets storage and loading for large multimodal trajectory collections \cite{Leal2024SARART, Chen2025RoboDM}. RoboCrowd shows that carefully designed interfaces can broaden participation in bimanual teleoperation data collection \cite{Mirchandani2025RoboCrowd}. These contributions make operational systems more scalable, but retain the assumption that the robot stack can already expose its intended sensing and control interfaces.

Robot failure-recovery research typically begins later in the stack. Self-Recovery Prompting revises prompts when a service robot encounters missing information, incorrect plans, or plan-execution failures \cite{Shirasaka2023SelfRecovery}; CoPAL replans when semantic or physical constraints make an action sequence infeasible \cite{Joublin2024CoPAL}; VLM-based recovery methods improve motion- and task-level corrections through optimized visual and textual prompts \cite{Chen2024VLMRecovery}; and RACER trains a language-guided visuomotor policy on augmented failure-recovery trajectories \cite{Dai2025RACER}. These methods recover behavior after an executable robot interface exists. SPINE addresses pre-operational faults that can prevent perception, middleware bringup, or teleoperation from becoming available in the first place.

\subsection{Language-guided robot agents}
Language and multimodal models provide increasingly general interfaces to robot behavior. SayCan grounds language-model proposals in learned affordances, Code as Policies converts instructions into executable programs, and PaLM-E and OpenVLA connect language and vision to embodied state or action \cite{Ahn2022, Liang2022CodeAsPolicies, Driess2023, Kim2024OpenVLA}. VLFM combines language semantics with geometric maps for zero-shot navigation on a physical robot \cite{Yokoyama2024VLFM}, while CoPAL and SELP add feedback or formal constraints to generated plans \cite{Joublin2024CoPAL, Wu2025SELP}. DROC instead distills online language corrections into a retrievable knowledge base, allowing later tasks to reuse prior human feedback \cite{Zha2024DROC}.

Closer to the systems boundary, ROS-LLM maps natural-language task requirements into ROS behaviors with environment and human feedback, while ROSA exposes ROS capabilities through constrained tools and natural-language interaction \cite{Mower2024ROSLLM, Royce2025ROSA}. Such interfaces lower the expertise needed to invoke an operational robot. SPINE complements them by recovering the device and middleware path beneath that interface. Its profile stores not only callable capabilities but also source contracts, live probes, hidden readiness checks, and operator-confirmed physical playbooks needed when invocation fails.

\subsection{Tool-using agents, memory, and validation}
Software-engineering agents demonstrate that language models can inspect repositories, invoke tools, revise code, and use execution feedback over multiple steps \cite{Yang2024, Jimenez2023, Wu2023, Shinn2023}. Their closure signals remain digital: a unit test passes, a build succeeds, or a repository issue is resolved. Robot operationalization introduces observations that source inspection cannot settle, including cable state, device enumeration, bus connectivity, controller readiness, and the outcome of a live robot-level probe.

Persistent context and validation therefore play different roles in SPINE. The category profile preserves platform contracts and executable diagnostics, while incident memory retains only fixes supported by a passed validation record. The runtime does not treat an LLM's diagnosis as resolution. It adjudicates parallel evidence, routes the next fix according to whether it is digitally or physically actionable, and reruns the complete goal-conditioned readiness bundle. This combines the iterative tool use of software agents with a cyber-physical closure condition.

\begin{figure*}[t]
\centering
\includegraphics[width=0.85\textwidth]{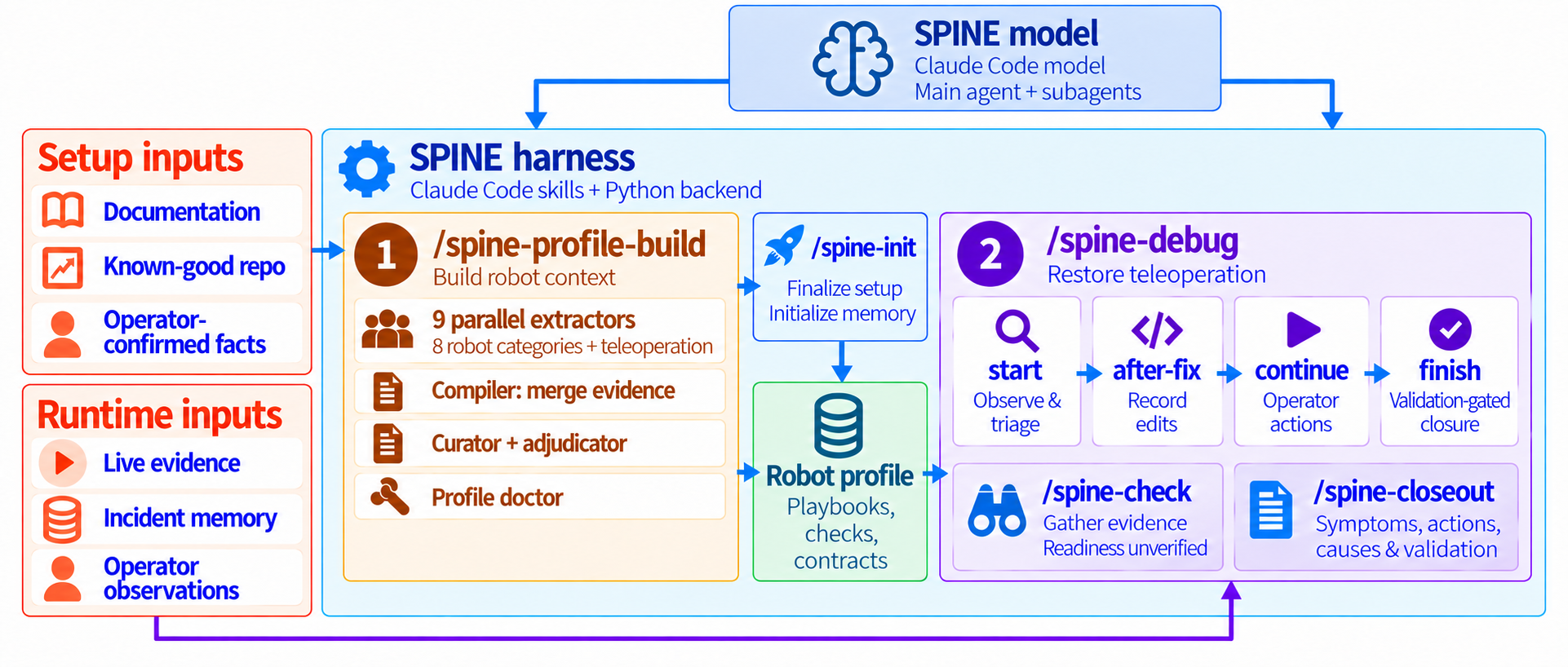}
\caption{SPINE system overview. The profile builder compiles robot-specific context, while the debugger combines this context with live evidence and incident memory to coordinate software repairs, guide physical interventions, and verify teleoperation readiness.}
\label{fig:system}
\end{figure*}

\section{SPINE System}\label{sec:spine_system}

SPINE combines language-model reasoning with an executable harness for restoring teleoperation. Two Claude Code skills organize the system: \texttt{/spine-profile-build} constructs robot-specific context, and \texttt{/spine-debug} uses that context during recovery. Their \texttt{SKILL.md} files define procedures, permitted tools, and delegation rules, while a Python backend manages execution, persistent state, and validation. The main agent selects interventions, but successful completion requires execution evidence checked by the backend.

Figure~\ref{fig:architecture} summarizes how SPINE establishes robot context and reuses it during debugging. For a new installation, the operator supplies documentation, a known-good software repository, and setup facts. The profile builder converts these inputs into a checked reference containing operating requirements, troubleshooting procedures, and validation criteria. The retained evidence, compiled profile, and incident history accumulated across sessions form the persistent robot context.

When teleoperation fails, the debugger combines this context with current observations in an observe--triage--repair--verify loop. It identifies active failures, coordinates diagnosis, and selects software repairs or guided physical interventions. Unresolved failures return to diagnosis with updated evidence; completion requires the live teleoperation pipeline and required readiness checks to pass after the latest intervention. The following subsections describe how the harness constructs this reference (Section~\ref{sec:profile_building}) and implements the debugging process (Section~\ref{sec:spine_debugging}).

\subsection{Robot Profile Building Skill}\label{sec:profile_building}

The \texttt{/spine-profile-build} skill combines parallel extraction, agent review, and deterministic checks to construct the robot reference shown in Fig.~\ref{fig:system}. A Python compiler first assembles an evidence bundle containing source text, configuration data, file references, and command relationships. Nine extraction subagents then work in parallel: eight cover robot identity, components, interfaces, software, configuration, procedures, safety, and diagnostics; the ninth reconstructs the teleoperation pipeline. Each receives instructions and an evidence packet and returns a JSON draft rather than modifying the final profile. The compiler merges these drafts with deterministic extraction while preserving source references.

The canonical profile consists of eight category JSON files and a manifest. Runtime adapters expose this information as \emph{contracts} describing operating requirements, \emph{probes and checks} that evaluate those requirements, and \emph{hardware playbooks} pairing physical interventions with verification procedures. A directed acyclic command graph specifies the teleoperation pipeline, including execution environments, dependencies, and success criteria. Its nodes distinguish one-shot setup commands, persistent bringup processes, and the final teleoperation command. Mutable values, such as camera identifiers and device paths, remain setup assumptions that subsequent observations can challenge.

Operator review connects the generated reference to the actual installation. The builder requests confirmation of the command graph, required interfaces, hardware playbooks, and individual hidden readiness checks. These checks target configuration or device problems that a short teleoperation run may miss; they contain no hidden fault annotations. The build also requires a recorded successful execution of the confirmed command graph.

A separate review stage checks the compiled profile against its evidence. Nine parallel curator subagents propose revisions, and one adjudicator resolves their proposals and cross-category conflicts. Python applies accepted changes and runs the \emph{profile doctor}, which checks schema consistency, evidence coverage, runtime compatibility, command-graph structure, operator confirmations, and build-time validation. The profile is sealed only after these requirements are satisfied. The supporting \texttt{/spine-init} skill wraps profile-first setup and initializes per-robot memory.

\subsection{Debugging Skill}\label{sec:spine_debugging}

The \texttt{/spine-debug} skill implements the runtime side of the SPINE harness (Fig.~\ref{fig:system}), combining the compiled robot profile with live evidence, incident memory, and operator observations. The profile supplies operating requirements, checks, and hardware playbooks. Language-model agents interpret evidence and select repairs, while the Python backend executes validation, maintains session state, and determines whether execution results permit successful closure.

The \texttt{debug start} command loads the profile, accepts supplied terminal output or collects fresh evidence through the declared teleoperation pipeline, and runs teleoperation-scoped hidden readiness checks. The backend maintains a \emph{blocker ledger} linking active failures to their evidence and originating checks. Each validation cycle refreshes this ledger, so subsequent decisions address current failures while retaining earlier observations as history.

When specialist diagnosis is required, four subagents examine terminal evidence, software and configuration deviations, hardware visibility, and readiness-check relevance in parallel. They return evidence-backed assessments without modifying the robot repository, directing the operator, or declaring success. The backend validates and aggregates their reports; the main agent selects the next intervention. This comparison helps distinguish shared symptoms: an unavailable camera stream may require correcting a configured identifier, restoring a physical connection, or both.

For software repair, the main agent edits the relevant code, configuration, or environment and calls \texttt{debug after-fix} to record the changes and rerun validation. Repairs must preserve required functionality rather than disable components or suppress failures. For physical repair, the agent presents one action from the affected component's hardware playbook. The operator's response enters through \texttt{debug continue}, which records the action and executes its designated probe or captures a required manual observation. Persistent step history guides subsequent instructions and avoids repeating confirmed actions. These targeted checks can advance troubleshooting but cannot establish overall readiness. Thus, \texttt{after-fix} and \texttt{continue} are alternative, repeatable intervention interfaces, not mandatory consecutive stages.

Full validation executes the confirmed teleoperation command graph with its dependencies, evaluates process output and declared success criteria, and cleans up the processes afterward. The final command has a profile-defined observation window, defaulting to 10 seconds. Hidden readiness checks supplement live execution with configuration and device checks. The \texttt{debug finish} command permits closure only when both validation results pass in the same latest cycle after the most recent recorded intervention, with no remaining blocking conditions. Earlier success cannot justify closure after further changes. Checks unavailable because of incomplete profile data are recorded as inconclusive and currently do not block closure; the readiness claim therefore covers only the executed checks and observation window.

The supporting \texttt{/spine-check} skill gathers focused evidence without replacing this completion gate. At closeout, \texttt{/spine-closeout} stores symptoms, attempted repairs, diagnoses, corrective actions, and validation evidence. Resolved records require a reference to passing validation; otherwise, they retain the diagnosis and remaining blocker. Incident memory guides future investigations, while fresh evidence remains necessary to establish the current cause and recovery.

\begin{table}[!t]
\captionsetup{font=footnotesize}
\centering
\caption{Injected debugging scenarios. D/P denote DOBOT/PiPER; SW, HW, and HY denote software-only, hardware-only, and hybrid faults.}
\label{tab:bug_cases}
\footnotesize
\setlength{\tabcolsep}{3pt}
\renewcommand{\arraystretch}{1.05}
\begin{tabularx}{\columnwidth}{@{}l >{\raggedright\arraybackslash}X@{}}
\toprule
\textbf{ID} & \textbf{Description} \\
\midrule
\multicolumn{2}{@{}l}{\textit{DOBOT X-Trainer}} \\
\addlinespace[2pt]

D-SW1 &
\textbf{Software:} We assign mismatched communication endpoints to the robot server and teleoperation client and alter launcher behavior to occupy a required local port. \\
\addlinespace[2pt]

D-SW2 &
\textbf{Software:} We alter the startup compatibility check to reject the installed controller or library version and assign a nonexistent serial number to one camera role. Packages and firmware remain unchanged. \\
\addlinespace[2pt]

D-HW1 &
\textbf{Hardware:} We latch the emergency-stop button, preventing arm actuation without changing software or connections. \\
\addlinespace[2pt]

D-HW2 &
\textbf{Hardware:} We unplug the right wrist camera cable and replace the left wrist camera cable with a known-broken cable. \\
\addlinespace[2pt]

D-HW3 &
\textbf{Hardware:} We disconnect one leader-arm cable and one follower-arm Ethernet cable, interrupting operator input and follower control. \\
\addlinespace[2pt]

D-HY1 &
\textbf{Hardware:} We replace the right wrist camera with a known-broken unit.
\textbf{Software:} We assign that camera role a nonexistent serial number. \\
\addlinespace[2pt]

D-HY2 &
\textbf{Hardware:} We unplug the robot-control Ethernet cable.
\textbf{Software:} We change both controller IP addresses in the launcher to the wrong subnet, leaving host network settings unchanged. \\
\midrule
\multicolumn{2}{@{}l}{\textit{AgileX PiPER}} \\
\addlinespace[2pt]

P-SW1 &
\textbf{Software:} We configure an incorrect USB bus address for the left CAN adapter and prioritize user-site Python packages over the intended environment, causing an incompatible Pinocchio import. \\
\addlinespace[2pt]

P-SW2 &
\textbf{Software:} We assign an incorrect serial number to the left wrist camera and change the teleoperation launch file's inverse-kinematics parameter path to a nonexistent file. \\
\addlinespace[2pt]

P-HW1 &
\textbf{Hardware:} We disconnect the left follower arm's CAN signal wire or connector and unplug the left wrist camera's USB cable. \\
\addlinespace[2pt]

P-HY1 &
\textbf{Hardware:} We unplug the left wrist camera's USB cable.
\textbf{Software:} We assign that camera an incorrect serial number in its launch file. \\
\addlinespace[2pt]

P-HY2 &
\textbf{Hardware:} We unplug the left USB-to-CAN adapter's USB connection.
\textbf{Software:} We increase the setup script's expected adapter count from two to three. \\
\bottomrule
\end{tabularx}
\end{table}

\section{Experimental Design}\label{sec:experimental_setup}

We evaluate SPINE by introducing a carefully selected set of software, hardware, and combined faults into otherwise functional teleoperation systems. An administrator prepares each faulty system without revealing the implanted faults to the participant. A robotics novice then attempts to restore teleoperation using SPINE, while human operators independently attempt the same scenarios using Claude Code and their own expertise.

\subsection{Robot Platforms}\label{sec:robot_platforms}

The DOBOT X-Trainer benchmark uses a bimanual teleoperation platform with two follower arms, two leader arms, RealSense cameras, a control workstation, and a vendor SDK over static Ethernet. The AgileX PiPER evaluation uses a ROS Noetic and CAN-based bimanual platform with four PiPER 6-DOF arms, USB-to-CAN adapters, and RealSense cameras. The two platforms differ in actuation bus, middleware, camera integration, and emergency-stop behavior, which allows us to test whether the same debugging structure transfers beyond one robot.

\subsection{Bug Cases}\label{sec:bug_cases}

We construct seven DOBOT scenarios and five PiPER scenarios across three categories: software-only faults alter runtime code or configuration; hardware-only faults change physical connections, components, or safety-control states; and hybrid faults combine software and physical changes. Each scenario contains two faults except D-HW1, which contains a single latched emergency stop. See Table~\ref{tab:bug_cases} for detailed descriptions of the faults introduced in each scenario. Table~\ref{tab:bug_cases} specifies the software mutations and physical interventions for every scenario.

We select these faults to cover device identification, communication, runtime compatibility, configuration, and physical connectivity. The compound scenarios test whether participants identify all required repairs rather than stop after correcting the first visible error. Hybrid cases further test whether participants distinguish software misconfiguration from physical failure when both affect the same subsystem. These scenarios provide targeted coverage of integration problems rather than an exhaustive or frequency-weighted sample of robot failures.

\subsection{Experimental Protocol}\label{sec:experimental_protocol}

We compare novice-operated SPINE with human operators on the same bug cases for both DOBOT X-Trainer and AgileX PiPER. An administrator prepares each trial from the reference software stack, implants the specified faults, and withholds the scenario identity and implantation records from participants.

In the SPINE condition, the agent uses a robot profile built from the working system before fault implantation. The novice initiates debugging, supplies requested observations, and follows SPINE's physical instructions, while SPINE directs diagnosis, software repair, and revalidation. In the human-operator condition, participants use standard Claude Code with Claude Sonnet~5.0 as the only permitted language model and independently choose diagnostic commands, software edits, and physical interventions. They receive the same underlying reference materials, but not SPINE's compiled profile, debugging skills, orchestration, or incident memory.

Both conditions aim to restore the standard teleoperation workflow. Successful recovery requires the operator to demonstrate that the follower arms and grippers respond to commands from the master arms while the required camera streams remain available; launching processes without errors does not suffice. SPINE additionally enforces its profile-declared validation pipeline and readiness checks before declaring completion.

\begin{table*}[!t]
\captionsetup{font=footnotesize}
\centering
\caption{Per-scenario outcomes. Scenario IDs encode platform and fault class, with D/P denoting DOBOT/PiPER and SW/HW/HY denoting software-only, hardware-only, and hybrid faults. OSS is reported as mean$\pm$SD over human-operator trials and as a decimal success score for SPINE. TTO is reported as mean$\pm$SD when repeated measurements are available. DOBOT human trials pool seven operators, and PiPER human trials pool nine operators.}
\label{tab:scenario_results}

\footnotesize
\setlength{\tabcolsep}{2.4pt}
\renewcommand{\arraystretch}{1.12}

\makebox[\textwidth][c]{%
\begin{tabular}{
@{}
l
p{0.34\textwidth}
c
cc
c
cc
@{}
}
\toprule
& & & \multicolumn{2}{c}{SPINE} & & \multicolumn{2}{c}{Human operators} \\
\cmidrule(lr){4-5}\cmidrule(lr){7-8}
ID & Description & $n$ & OSS & TTO & $n$ & OSS & TTO \\
\midrule

\multicolumn{8}{@{}l}{\textit{DOBOT X-Trainer}}\\
D-SW1 & Mismatched local endpoint and busy port
& 3 & 1.00\,$\pm$\,0.00 & 6:20\,$\pm$\,0:01 & 7 & 0.36\,$\pm$\,0.38 & 7:28\,$\pm$\,3:05 \\

D-SW2 & Stale environment, version gate, camera ID
& 3 & 1.00\,$\pm$\,0.00 & 6:41\,$\pm$\,1:19 & 7 & 0.57\,$\pm$\,0.45 & 20:11\,$\pm$\,13:25 \\

D-HW1 & Latched emergency stop
& 3 & 1.00\,$\pm$\,0.00 & 13:35\,$\pm$\,0:13 & 7 & 1.00\,$\pm$\,0.00 & 20:46\,$\pm$\,6:12 \\

D-HW2 & Broken wrist-camera cabling
& 3 & 1.00\,$\pm$\,0.00 & 9:59\,$\pm$\,0:34 & 7 & 1.00\,$\pm$\,0.00 & 17:54\,$\pm$\,17:52 \\

D-HW3 & Leader and follower links unplugged
& 3 & 1.00\,$\pm$\,0.00 & 11:18\,$\pm$\,1:44 & 7 & 1.00\,$\pm$\,0.00 & 13:01\,$\pm$\,7:47 \\

D-HY1 & Broken wrist camera plus stale ID
& 3 & 1.00\,$\pm$\,0.00 & 14:55\,$\pm$\,0:55 & 7 & 0.64\,$\pm$\,0.38 & 11:55\,$\pm$\,7:44 \\

D-HY2 & LAN unplug plus wrong subnet
& 3 & 1.00\,$\pm$\,0.00 & 9:34\,$\pm$\,0:32 & 7 & 0.71\,$\pm$\,0.27 & 12:43\,$\pm$\,9:14 \\

\textbf{Overall} & &
& \textbf{1.00\,$\pm$\,0.00} & \textbf{10:20\,$\pm$\,3:11} &
& \textbf{0.76\,$\pm$\,0.36} & \textbf{14:51\,$\pm$\,10:44} \\

\midrule
\multicolumn{8}{@{}l}{\textit{AgileX PiPER}}\\

P-SW1 & CAN binding and Python path drift
& 3 & 1.00\,$\pm$\,0.00 & 11:32\,$\pm$\,0:11 & 9 & 0.94\,$\pm$\,0.17 & 16:00\,$\pm$\,10:16 \\

P-SW2 & Camera ID and IK config drift
& 3 & 1.00\,$\pm$\,0.00 & 8:28\,$\pm$\,0:06 & 9 & 1.00\,$\pm$\,0.00 & 10:56\,$\pm$\,13:01 \\

P-HW1 & Camera and CAN links unplugged
& 3 & 1.00\,$\pm$\,0.00 & 2:45\,$\pm$\,0:22 & 9 & 1.00\,$\pm$\,0.00 & 13:36\,$\pm$\,13:23 \\

P-HY1 & Camera unplug plus stale ID
& 3 & 1.00\,$\pm$\,0.00 & 10:44\,$\pm$\,0:19 & 9 & 1.00\,$\pm$\,0.00 & 13:43\,$\pm$\,8:07 \\

P-HY2 & CAN adapter unplug plus count drift
& 3 & 1.00\,$\pm$\,0.00 & 5:33\,$\pm$\,0:04 & 9 & 1.00\,$\pm$\,0.00 & 8:14\,$\pm$\,7:27 \\

\textbf{Overall} & &
& \textbf{1.00\,$\pm$\,0.00} & \textbf{7:48\,$\pm$\,3:27} &
& \textbf{0.99\,$\pm$\,0.07} & \textbf{12:30\,$\pm$\,10:35} \\

\bottomrule
\end{tabular}%
}
\end{table*}

\section{Results}\label{sec:results}

We evaluate fault resolution and debugging efficiency using the per-scenario outcomes in Table~\ref{tab:scenario_results}. OSS measures the fraction of implanted faults resolved within each trial, rather than the fraction of trials that achieve complete recovery. Timing statistics include unsuccessful trials, so we interpret them alongside OSS.

\subsection{DOBOT X-Trainer}

SPINE improved fault resolution and reduced mean time-to-teleoperation across the seven DOBOT scenarios. SPINE achieved an OSS of 1.00 in every scenario, compared with an overall human-operator score of 0.76. Mean time-to-teleoperation decreased from 14~min~51~s to 10~min~20~s, a 30\% reduction. The success advantage arose entirely from software-only and hybrid scenarios: human operators achieved mean OSS values of 0.36 and 0.57 on D-SW1 and D-SW2, and 0.64 and 0.71 on D-HY1 and D-HY2. Both conditions resolved all faults in the three hardware-only scenarios.

SPINE reduced mean time-to-teleoperation in six of seven scenarios, but its advantage did not always combine faster execution with more complete recovery. On D-HY1, which combined a broken camera with an incorrect camera identifier, SPINE took 14~min~55~s compared with 11~min~55~s for human operators, while achieving complete fault resolution rather than a mean OSS of 0.64. This exception highlights why shorter debugging duration alone does not establish better recovery performance.

\subsection{AgileX PiPER}

SPINE's principal advantage on PiPER was faster recovery while maintaining complete fault resolution. Across five scenarios, SPINE achieved an overall OSS of 1.00 compared with 0.99 for human operators, and reduced mean time-to-teleoperation from 12~min~30~s to 7~min~48~s, a 38\% reduction. SPINE recorded a lower mean time in every scenario. The largest reduction occurred on P-HW1, which combined disconnected camera and CAN links: mean time fell from 13~min~36~s to 2~min~45~s, approximately 80\%. Human operators already achieved complete fault resolution in four scenarios; their remaining unresolved faults occurred in P-SW1. These results extend the observed efficiency advantage to a second platform with a different middleware and communication stack, while showing only a small absolute improvement in success.

\subsection{Ablation Study}

Removing the robot profile reduced overall fault resolution and increased average debugging duration on DOBOT (Table~\ref{tab:dobot_ablation_results}). Mean OSS fell from 1.00 to 0.79, while mean duration increased from 10~min~20~s to approximately 17~min. The profile-free variant resolved only half the faults in D-HW2 and none in D-HY1. Although it completed several other scenarios faster than full SPINE, these local improvements did not offset its incomplete recovery and longer durations elsewhere.

Removing the hardware playbook produced a larger aggregate performance loss. Mean OSS fell to 0.57, and mean duration increased to 31~min~25~s, approximately three times that of full SPINE. This variant resolved no faults in D-HW1 or D-HW2 and only half the faults in each hybrid scenario. Both ablated variants retained complete fault resolution on the software-only scenarios, concentrating their observed failures in cases that required physical intervention.

The ablations provide preliminary evidence that robot-specific context and structured physical troubleshooting contribute to reliable recovery. However, each ablated condition includes only one trial per scenario, so these observations do not establish statistically reliable effect sizes or isolate the mechanisms behind individual failures.

\begin{table}[H]
\captionsetup{font=footnotesize}
\centering
\caption{DOBOT ablation study outcomes comparing full SPINE against two ablated variants. SPINE results report mean OSS and mean TTO over $n=3$ trials per scenario, while each ablation result reports a single trial ($n=1$).}
\label{tab:dobot_ablation_results}

\footnotesize
\setlength{\tabcolsep}{3.0pt}
\renewcommand{\arraystretch}{1.08}
\setlength{\dashlinedash}{1.2pt}
\setlength{\dashlinegap}{1.6pt}
\setlength{\arrayrulewidth}{0.35pt}

\begin{tabular}{@{}l|cc:cc:cc@{}}
\toprule
& \multicolumn{2}{c:}{SPINE} & \multicolumn{2}{c:}{No Profile} & \multicolumn{2}{c}{\shortstack{No Hardware\\Playbook}} \\
\cmidrule(lr){2-3}\cmidrule(lr){4-5}\cmidrule(lr){6-7}
ID & OSS & TTO & OSS & TTO & OSS & TTO \\
\midrule
D-SW1 & 1.00 & 6:20  & 1.00 & 4:40  & 1.00 & 20:47 \\
D-SW2 & 1.00 & 6:41  & 1.00 & 6:37  & 1.00 & 5:23  \\
D-HW1 & 1.00 & 13:35 & 1.00 & 11:01 & 0.00 & 45:00 \\
D-HW2 & 1.00 & 9:59  & 0.50 & 45:00 & 0.00 & 45:00 \\
D-HW3 & 1.00 & 11:18 & 1.00 & 23:52 & 1.00 & 32:55 \\
D-HY1 & 1.00 & 14:55 & 0.00 & 20:37 & 0.50 & 25:50 \\
D-HY2 & 1.00 & 9:34  & 1.00 & 7:25  & 0.50 & 45:00 \\
\midrule
\textbf{Overall} & \textbf{1.00} & \textbf{10:20} & \textbf{0.79} & \textbf{17:00} & \textbf{0.57} & \textbf{31:25} \\
\bottomrule
\end{tabular}
\end{table}

\section{Discussion}\label{sec:discussion}

Our results suggest that structured agentic assistance can improve robot debugging even when human operators already have access to a capable coding agent. The benefits differ across platforms: on DOBOT, SPINE improved fault resolution primarily in software-only and hybrid scenarios; on PiPER, where human operators already resolved nearly all faults, its principal advantage was faster recovery. These findings highlight two distinct requirements for effective debugging: identifying all faults that prevent operation and resolving them efficiently. The comparison evaluates the complete human--agent workflow, not the language model's reasoning ability in isolation.

The ablations suggest that robot-specific context and physical troubleshooting guidance help connect software diagnosis to hardware recovery. Both ablated variants retained complete fault resolution on software-only scenarios but left faults unresolved in hardware-only and hybrid cases. This pattern is consistent with the intended roles of the profile and hardware playbooks: the former establishes platform-specific expectations, while the latter structures interventions that require physical access. However, one trial per scenario for each ablation provides only preliminary evidence and cannot establish reliable effect sizes or explain individual failures conclusively.

Our experimental design limits the breadth of these conclusions. We evaluate controlled faults on two bimanual platforms, with few SPINE repetitions, rather than naturally occurring failures across diverse deployments. Furthermore, debugging durations include unsuccessful trials, so shorter times do not necessarily indicate faster successful recovery. The timing comparisons also begin after profile construction and therefore do not measure total setup effort. Larger studies should include additional embodiments, more operators, naturally occurring faults, and the cost of building and maintaining robot profiles.

SPINE's readiness claim also depends on the quality and coverage of its reference information and validation procedures. A passing teleoperation run and readiness checks establish operation only within the configured checks and observation window; they do not establish long-term reliability, autonomous policy readiness, or comprehensive safety. SPINE still requires operator review and human access for physical interventions. Future work should examine incomplete or outdated profiles and failures that emerge during sustained operation, while preserving the distinction between observed recovery and broader operational guarantees.

\section{Conclusion}\label{sec:conclusion}

SPINE treats teleoperation readiness as a cyber-physical debugging problem that requires coordinated software repair, physical intervention, and live verification. Its profile builder and debugger turn robot-specific evidence into a reusable recovery workflow. Across two bimanual platforms, novice-operated SPINE resolved all implanted faults and reduced mean time-to-teleoperation relative to human operators using Claude Code. Within the evaluated scope, these results show how agentic AI can help bridge the gap between capable robot intelligence and the functioning physical systems it requires, making teleoperation recovery more accessible and efficient.

\section*{Acknowledgements}
We thank Guo Ye, Jihai Zhao, Tony Dong, Jiale Chen, Vincent Cheng, Soyeon Park, Dennis Wu, Hongyu Chen, Donggyun Lee, and Mingxian Wang, who volunteered as human operators in our experiments and/or ran experiments for the SPINE agent. We also thank Guo Ye for the numerous conversations about the design of SPINE, which helped inspire certain elements of it. AI tools (Claude and ChatGPT) helped us generate and refine the figures in this paper.

\end{document}